\documentclass[letterpaper,10pt,conference]{ieeeconf}

\IEEEoverridecommandlockouts
\usepackage{graphicx}
\usepackage{latexml}
\iflatexml\else\usepackage{capt-of}\fi
\usepackage{cite}
\usepackage{amsmath}
\usepackage{amssymb}
\usepackage{array}
\usepackage{booktabs}
\usepackage{url}
\usepackage{balance}
\title{\LARGE \bf
MUSBRIC: LEGO-Compatible Pneumatic Muscle Bricks for Reconfigurable Musculoskeletal Robots
}

\author{Qiyao Wang$^{1}$$^{\dagger}$, Yinglei Zhu$^{1}$$^{\dagger}$, Liang Zhou$^{1}$, Wenzhe Gong$^{2}$, Xin-Jun Liu$^{1}$ and Huichan Zhao$^{1*}$%
\thanks{$^{\dagger}$Contributed equally to this work.}%
\thanks{$^{1}$Department of Mechanical Engineering, State Key Laboratory of Tribology in Advanced Equipment, Beijing Key Laboratory of Transformative High-end Manufacturing Equipment and Technology, Tsinghua University, Beijing 100084, China.}%
\thanks{$^{2}$Tsinghua Medicine, Tsinghua University, Beijing 100084, China.}%
\thanks{$^{*}$Corresponding Author}%
}

\iflatexml
\else
\IEEEaftertitletext{%
  \begin{minipage}{\textwidth}
    \centering
    \includegraphics[width=\textwidth]{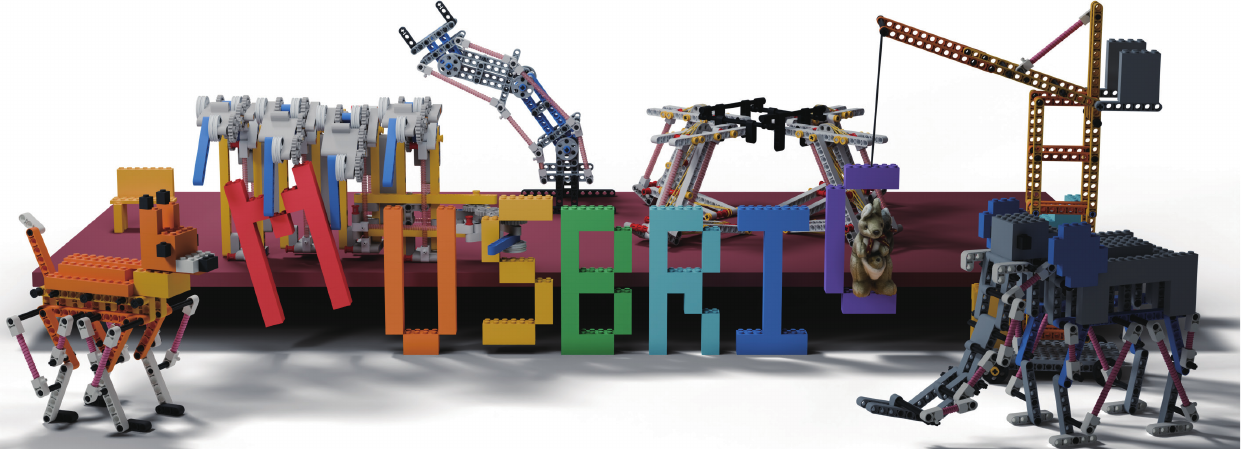}
    \captionof{figure}{MUSBRIC combines LEGO-compatible pneumatic muscle bricks and a corresponding simulation platform.}
    \label{fig:intro}
  \end{minipage}%
}
\fi

\begin{document}
\raggedbottom

\maketitle
\thispagestyle{empty}
\pagestyle{empty}

\begin{abstract}
Musculoskeletal robots combine compliant actuators with articulated skeletons, but exploring alternative configurations often requires repeated fabrication. We present MUSBRIC, a modular hardware platform and accompanying MuJoCo-based simulator for reconfigurable musculoskeletal robots. LEGO-compatible endcaps turn fiber-reinforced extensile pneumatic artificial muscles into reusable, plug-and-play muscle bricks that connect directly to standard LEGO components. Experimentally identified actuator behavior supports corresponding musculoskeletal simulations. An elbow-joint prototype validates load-dependent flexion and extension, showing that pulley-guided routing enables reversible motion and a mechanical stop stabilizes the straight posture. A robotic hand demonstrates compliant grasping and multi-joint construction. Matched hardware and simulation sweeps across 15 leg configurations demonstrate systematic exploration of attachment geometry, while a quadruped assembled from four legs demonstrates walking. Together, these demonstrations span mechanism analysis, configuration exploration, and functional robot construction within a single reconfigurable system. The hardware designs and simulation code will be released as open source soon to facilitate reproduction and further development.
\end{abstract}

\iflatexml
\begin{figure*}[t]
  \centering
  \includegraphics[width=\textwidth]{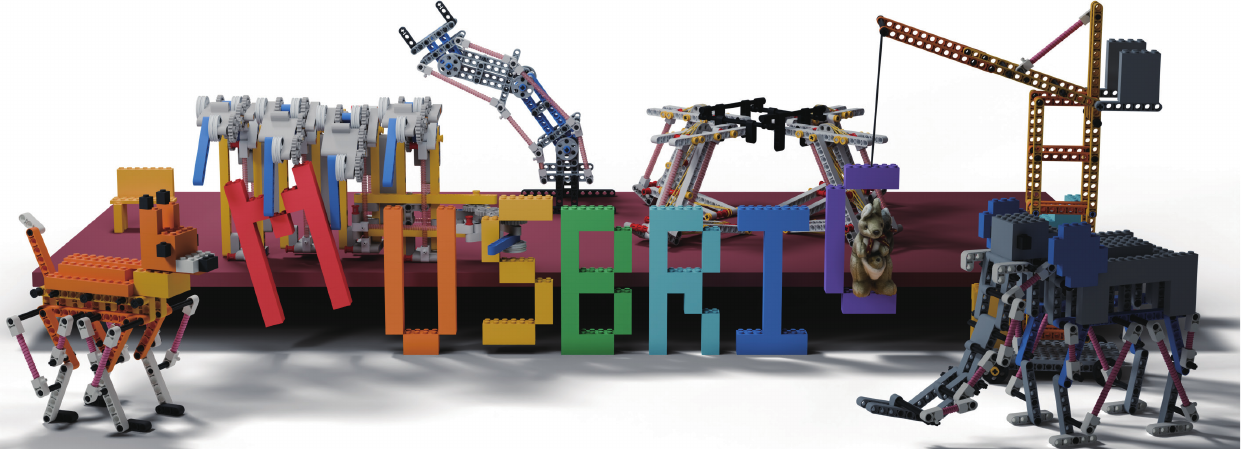}
  \caption{MUSBRIC combines LEGO-compatible pneumatic muscle bricks and a corresponding simulation platform.}
  \label{fig:intro}
\end{figure*}
\fi

\section{Introduction}
\label{sec:introduction}

Rigid--soft hybrid robots combine structural support with compliant actuation to achieve adaptable and physically robust behavior~\cite{rus2015softrobots,zhang2019muscles}. In musculoskeletal implementations, artificial muscles provide compliant force generation, while articulated skeletons define joint axes and motion patterns. This architecture has been explored in soft--rigid hybrid manipulators~\cite{tondu2005seven,bruder2023payload}, bio-inspired legged robots~\cite{dong2025bionicleg,wang2013quadruped,asano2017humanoid}, and robotic hands and grasp-assist gloves~\cite{deimel2016hand,in2015exoglove}. Their behavior depends strongly on the mechanical configuration: muscle routing, link geometry, and joint constraints together determine torque transmission, equilibrium states, and the reachable workspace. Modeling frameworks now cover musculoskeletal, embedded-skeleton, and hybrid rigid--soft systems~\cite{zhang2019musculoskeletal,bern2022skeletons,mathew2023sorosim}, while computational methods support automated morphology and actuator design~\cite{hiller2012design,connolly2017automatic}. Nevertheless, comparing alternative musculoskeletal configurations in hardware remains costly when each change requires custom fabrication and assembly.

Modular soft robotics addresses this construction problem by reusing components across assemblies. Prior systems use detachable hard--soft magnetic interfaces~\cite{kwok2014magnetic}, standardized pneumatic blocks~\cite{lee2016sobl}, reconfigurable multi-material voxels~\cite{legrand2023voxel}, modular vacuum muscles~\cite{lee2024rovpam}, and LEGO-compatible pneumatic components~\cite{lee2018softlego} to enable reusable construction and reconfiguration of soft robots. Musculoskeletal robots, however, require the coordinated configuration of muscle attachments, skeletal geometry, joints, and transmissions. This broader design space calls for a more comprehensive modular system that covers both compliant actuation and rigid structures, together with simulation models for evaluating alternative assemblies.

We introduce MUSBRIC, a LEGO-compatible platform that unifies compliant muscle bricks, rigid skeletons, and a matched simulator for building and reconfiguring musculoskeletal robots (Fig.~\ref{fig:intro}). MUSBRIC uses fiber-reinforced extensile pneumatic artificial muscles (PAMs)~\cite{hawkes2016ipam,lin2022anthropomorphic} as compliant actuators and LEGO Technic\footnote{LEGO Technic is a trademark of the LEGO Group, which does not sponsor, authorize, or endorse this work.} components as articulated skeletons, connected through custom endcaps compatible with LEGO pins and axles. Pressure-dependent force--strain measurements identify the actuator parameters, which are incorporated into a simulator built on EquiMus~\cite{zhu2025equimus} and MuJoCo~\cite{todorov2012mujoco}. We validate MUSBRIC at increasing levels of complexity: an antagonistic elbow joint reveals how actuator routing and joint constraints govern reversible motion and posture stability; a robotic hand achieves compliant multi-joint grasping; and matched simulation and hardware sweeps across 15 leg configurations explore attachment geometry before four legs are assembled into a walking quadruped.

The main contributions of this work are:
\begin{enumerate}
  \item LEGO-compatible muscle bricks for constructing reconfigurable musculoskeletal robots.
  \item A MuJoCo-based simulator extending EquiMus with identified actuator models for reconfigurable musculoskeletal assemblies.
  \item Validation across musculoskeletal mechanisms and robots, showing MUSBRIC's utility for mechanical analysis, rapid design exploration, and functional tasks.
\end{enumerate}
All hardware designs and simulation code will be released as open source soon.

\section{Hardware Design}
\label{sec:hardware}

\begin{figure*}[!t]
  \centering
  \includegraphics[width=\textwidth]{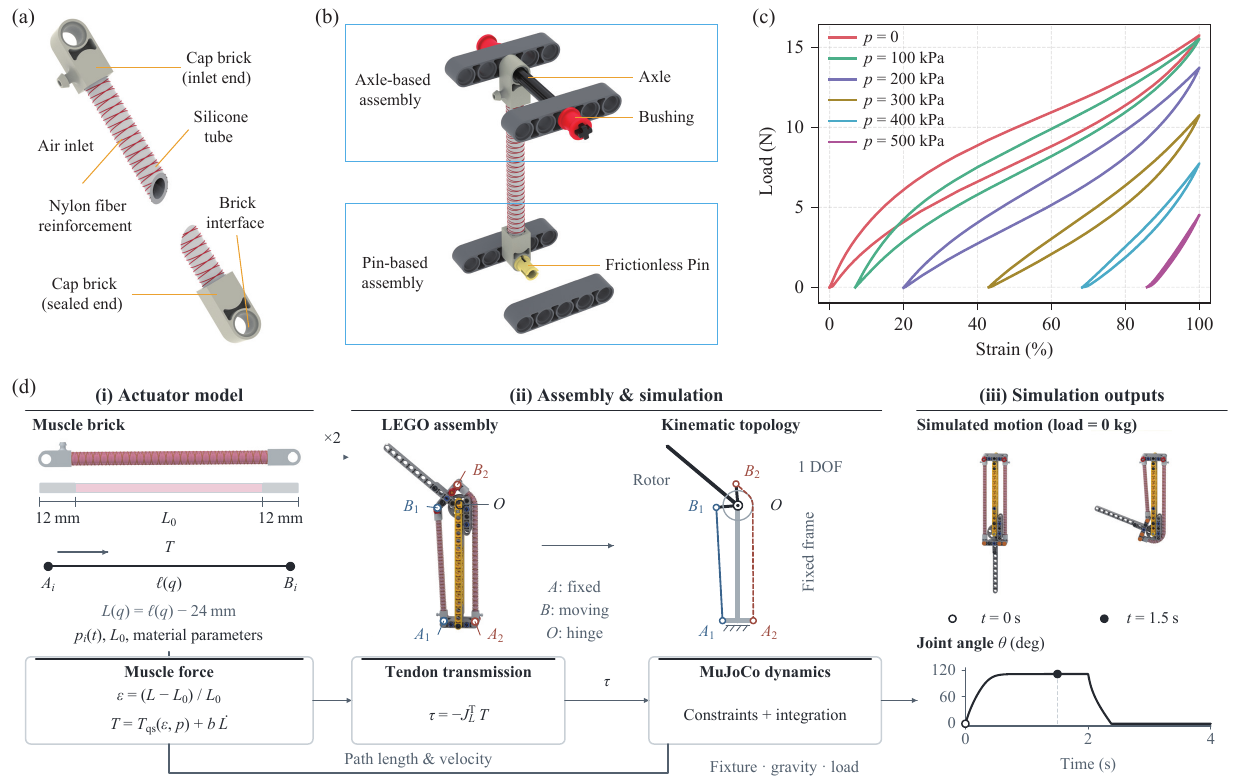}
  \caption{Muscle-brick design and modeling. (a) Structure of the muscle brick, including the fiber-reinforced silicone tube and the inlet and sealed cap bricks. (b) Connection to LEGO beams using an axle and bushing (top) or a frictionless pin (bottom). (c) Tensile load--strain curves of an actuator at 0--500~kPa. (d) Simulation method coupling the actuator model with assembly dynamics in MuJoCo.}
  \label{fig:design-and-sim}
\end{figure*}

\subsection{LEGO-Compatible Muscle Bricks}
\label{subsec:muscle-bricks}

The muscle brick is built around a fiber-reinforced extensile PAM, which uses a silicone tube as its body and wound nylon fibers as reinforcement (Fig.~\ref{fig:design-and-sim}a). The nylon fibers are wound at a small angle to the tube's cross-sectional plane, strongly constraining radial expansion while leaving axial deformation comparatively unrestricted~\cite{hawkes2016ipam,lin2022anthropomorphic}. The pressurized PAM therefore elongates with little change in outer diameter. Compared with the radial expansion of a McKibben muscle, this nearly constant profile reduces interference with surrounding LEGO structures. Two custom endcaps terminate the actuator: one admits compressed air, while the other seals the tube.

To fabricate each actuator, a silicone tube is wound with nylon fiber at controlled pitch and tension using a motorized winding stage. The wound tube is then coated with Sil-Poxy to secure the fibers. After curing, the reinforced tube is cut to the required length and bonded to the two custom endcaps.

\subsection{Assembly Interfaces}
\label{subsec:assembly}

The custom endcaps use a LEGO-compatible hole that accepts either a pin or an axle, allowing the actuator to connect with standard beams, axles, bushings, gears, and other structural and transmission components (Fig.~\ref{fig:design-and-sim}b). A pin joins the endcap to a beam hole while allowing it to rotate; alternatively, an axle passes through the same hole and can carry bushings or gears. The same actuator can therefore be repositioned or integrated into different assemblies without modification.

\section{Modeling and Simulation}
\label{sec:simulation}

\subsection{Constitutive Characterization}
\label{subsec:characterization}

The axial behavior of an actuator is described by three quantities: internal pressure $p$, axial tensile load $T$, and axial strain $\varepsilon$. Let $\ell$ be the attachment-to-attachment path length and $L=\ell-2L_c$ the pneumatic length, where $L_c=12~\mathrm{mm}$ is the rigid length at each end. With pneumatic rest length $L_0$, $\varepsilon=(L-L_0)/L_0$. Under quasi-static conditions, the elastic restoring force of the reinforced tube, $F_{\mathrm{elas}}(\varepsilon)$, balances the applied tensile load and the pressure-induced axial force:
\begin{equation}
  F_{\mathrm{elas}}(\varepsilon)=T+pS_{\mathrm{eff}}(p).
  \label{eq:equilibrium}
\end{equation}
where $S_{\mathrm{eff}}(p)$ is the effective pressure area over which the internal pressure produces an axial force. The nylon fibers constrain radial expansion, while pressurization deforms the silicone wall and changes the lumen area. This pressure dependence is approximated to first order as
\begin{equation}
  S_{\mathrm{eff}}(p)=S_0+\alpha p.
  \label{eq:effective-area}
\end{equation}
Combining the tensile-load and pressure-induced terms defines the effective force
\begin{equation}
  F_{\mathrm{eff}}=T+p(S_0+\alpha p),
  \label{eq:effective-force}
\end{equation}

Uniaxial tensile tests were conducted at 0--500~kPa in 100~kPa increments and a crosshead speed of 100~mm/min. At each pressure, the actuator was quasistatically stretched to 100\% strain and then unloaded to its initial length. Two preconditioning cycles were first applied to reduce interference from the Mullins effect and stabilize the material response; data from the third cycle were recorded. Figure~\ref{fig:design-and-sim}c shows the recorded loading--unloading cycle at each pressure. The separation between the two branches indicates hysteresis.

To obtain a constitutive relation for simulation, all pressure groups are expressed in the effective-force coordinate and fitted jointly. The resulting cubic relation is
\begin{equation}
  \widehat{F}_{\mathrm{eff}}(\varepsilon)
  =\sum_{j=0}^{3}c_j\varepsilon^j.
  \label{eq:constitutive}
\end{equation}
The identified parameters are listed in Table~\ref{tab:constitutive}. The fitted relation is a common quasi-static approximation over the calibrated range and does not separately model hysteresis. For simulation, the quasi-static actuator tension is therefore calculated as
\begin{equation}
  T_{\mathrm{qs}}(\varepsilon,p)
  =\widehat{F}_{\mathrm{eff}}(\varepsilon)-p(S_0+\alpha p)
  =\sum_{j=0}^{3}c_j\varepsilon^j-p(S_0+\alpha p).
  \label{eq:quasistatic-force}
\end{equation}
To represent the dynamic response in simulation, the quasi-static relation is augmented with a linear axial viscous term:
\begin{equation}
  T_i=T_{\mathrm{qs}}(\varepsilon_i,p_i)+b\dot L_i.
  \label{eq:damped-force}
\end{equation}
The damping coefficient $b$ is subsequently identified from the experiments in Section~\ref{subsec:joint-validation}.

\begin{table}[!t]
  \caption{Identified parameters of the PAM constitutive model.}
  \label{tab:constitutive}
  \centering
  \small
  \setlength{\tabcolsep}{0.06in}
  \renewcommand{\arraystretch}{1.06}
  \begin{tabular}{l c}
    \toprule
    Quantity & Identified value \\
    \midrule
    $S_0$ & $12.56637~\mathrm{mm}^2$ \\
    $\alpha$ & $3.03153\times10^{-11}~\mathrm{m}^2/\mathrm{Pa}$ \\
    $c_0$ & $0.095067~\mathrm{N}$ \\
    $c_1$ & $31.463491~\mathrm{N}$ \\
    $c_2$ & $-44.708056~\mathrm{N}$ \\
    $c_3$ & $30.779766~\mathrm{N}$ \\
    \bottomrule
  \end{tabular}
\end{table}

\subsection{Simulation Framework}
\label{subsec:equimus}

Figure~\ref{fig:design-and-sim}d summarizes the simulation workflow. At each time step, the assembly topology provides the actuator path lengths $\boldsymbol{\ell}(\mathbf q)$. For actuator $i$, $L_i=\ell_i-2L_c$; $\varepsilon_i$, $\dot L_i$, and $p_i$ then determine $T_i$ through Eq.~(\ref{eq:damped-force}). The tensions are mapped to generalized torques by
\begin{equation}
  \boldsymbol{\tau}_{\mathrm{act}}
  =-\mathbf J_\ell(\mathbf q)^{\mathsf T}\mathbf T,
  \qquad
  \mathbf J_\ell=\frac{\partial\boldsymbol{\ell}}{\partial\mathbf q},
  \label{eq:virtual-work}
\end{equation}
where $\mathbf q$ denotes the generalized coordinates, $\boldsymbol{\ell}$ and $\mathbf T$ collect the actuator path lengths and tensions, respectively, $\mathbf J_\ell$ is the path-length Jacobian, and $\boldsymbol{\tau}_{\mathrm{act}}$ is the resulting generalized torque. MuJoCo then integrates the constrained dynamics to obtain the assembly motion $\mathbf q(t)$.

\section{Elbow-Joint Mechanics}
\label{sec:joint}

\begin{figure*}[!t]
  \centering
  \includegraphics[width=\textwidth]{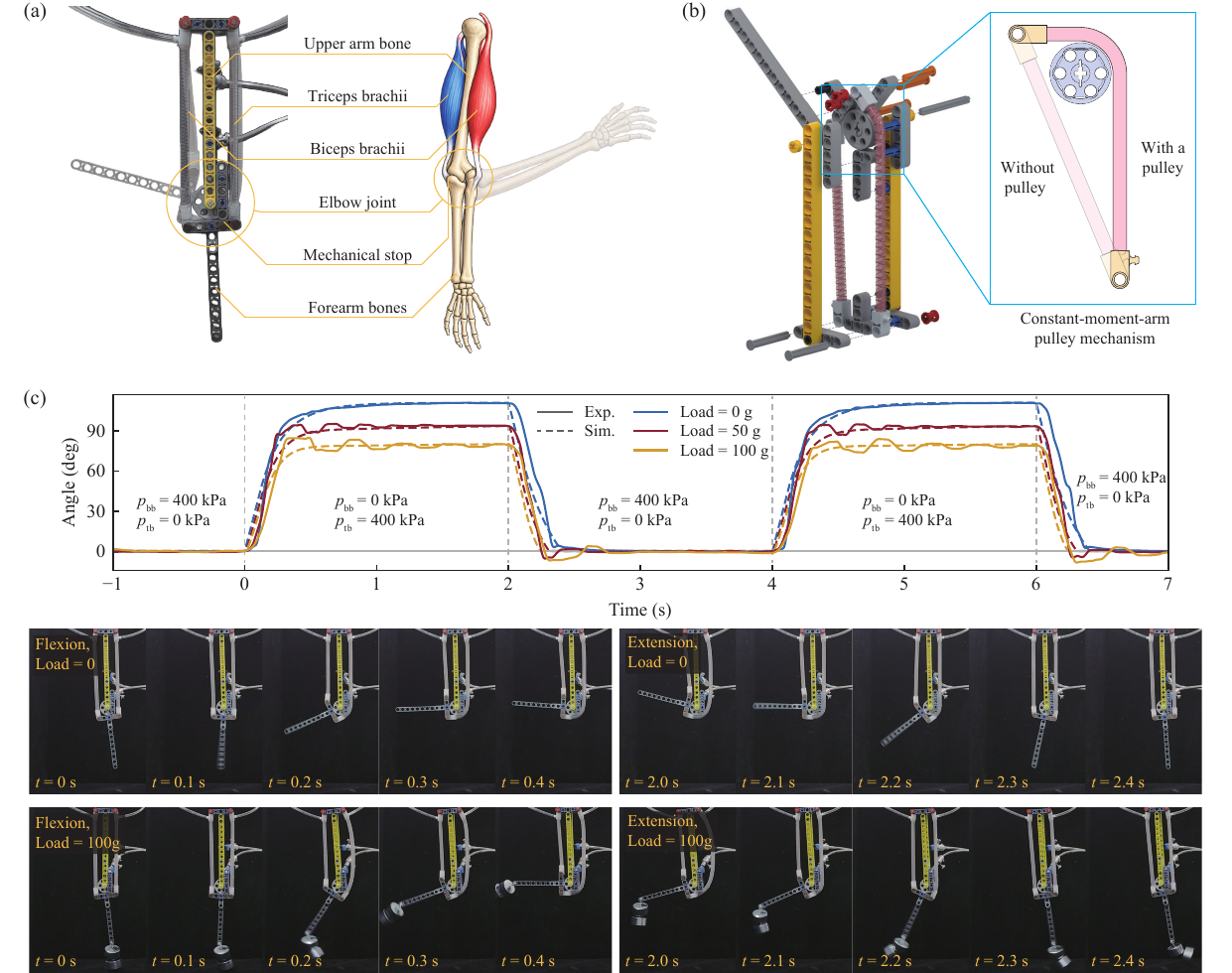}
  \caption{Single-joint mechanism and load response. (a) Elbow joint constructed using MUSBRIC. (b) Pulley-guided routing and comparison with direct attachment. (c) Experimental (solid) and simulated (dashed) joint angles under 0, 50, and 100~g added loads. The image sequences show flexion and extension at 0 and 100~g. Pressure commands alternate between $(p_{\mathrm{bb}},p_{\mathrm{tb}})=(400,0)$ and $(0,400)$~kPa.}
  \label{fig:joint}
\end{figure*}

\subsection{Bioinspired Layout}
\label{subsec:joint-layout}

The single-joint prototype uses two antagonistic actuators to rotate a LEGO forearm about a fixed upper segment (Fig.~\ref{fig:joint}a). The actuators occupy the biceps brachii and triceps brachii positions, with pressures denoted by $p_{\mathrm{bb}}$ and $p_{\mathrm{tb}}$. The joint angle $q$ is zero when the two segments are straight and positive during flexion. A mechanical stop blocks rotation beyond the straight configuration, allowing the joint to bend only in the flexion direction.

Actuator routing determines the torque transmitted to the joint. For actuator $i$ with length $\ell_i$ and tension $T_i$, the signed moment arm is
\begin{equation}
  r_i=-\frac{\partial\ell_i}{\partial q}.
\end{equation}
The corresponding joint torque is $r_iT_i$. Figure~\ref{fig:joint}b compares direct and pulley-guided routing of the triceps actuator. Without the pulley, the triceps moment arm $r_{\mathrm{tb}}$ decreases as $q$ increases and can change sign, whereas the biceps moment arm $r_{\mathrm{bb}}$ remains positive. Once $r_{\mathrm{tb}}$ changes sign, the triceps actuator reverses its contribution to the joint torque, and antagonistic pressure switching can no longer return the joint from the resulting high-angle region. With the pulley, $r_{\mathrm{tb}}$ remains nearly constant over the operating range, avoiding this sign reversal and supporting consistent, reversible motion~\cite{lin2022anthropomorphic}. For an ideal pulley of radius $r_p$, the triceps actuator path satisfies $\ell_{\mathrm{tb}}(q)=\ell_{\mathrm{tb}}(0)+ r_pq$ and hence $|r_{\mathrm{tb}}|=r_p$ over the wrapped range. The biceps moment arm $r_{\mathrm{bb}}$ is not constant but varies only slightly during the bending.

\subsection{Pressure-Step Response}
\label{subsec:joint-validation}

To characterize the joint response to pressure steps, the antagonistic actuators are driven by an alternating open-loop command sequence. The command alternates every 2~s between the flexion state, $(p_{\mathrm{bb}},p_{\mathrm{tb}})=(0,400)$~kPa, and the extension state, $(400,0)$~kPa. The same sequence is applied with added masses of 0, 50, and 100~g attached to the distal end of the forearm. Joint angles are measured from video by fitting the LEGO hole centers along each link. Figure~\ref{fig:joint}c shows the repeated flexion--extension responses to these pressure steps. Under the flexion command, the settled angles are $110.6^\circ$, $93.6^\circ$, and $78.7^\circ$ for added loads of 0, 50, and 100~g, respectively.

\begin{table}[!t]
  \caption{Experimental and simulated joint responses.}
  \label{tab:joint}
  \centering
  \small
  \setlength{\tabcolsep}{0.06in}
  \renewcommand{\arraystretch}{1.06}
  \begin{tabular}{c c c c}
    \toprule
    Added load & \multicolumn{2}{c}{Settled angle ($^\circ$)} & Trace RMSE \\
    (g) & Experiment & Simulation & ($^\circ$) \\
    \midrule
    0   & 110.6 & 110.9 & 7.09 \\
    50  & 93.6  & 92.8  & 5.26 \\
    100 & 78.7  & 79.7  & 5.95 \\
    \bottomrule
  \end{tabular}
\end{table}

The damping coefficient $b$ in Eq.~(\ref{eq:damped-force}) is identified from the pressure-step responses under added loads of 0, 50, and 100~g, with one value shared by both actuators. The quasi-static coefficients, mass properties, and pressure commands are held fixed, while joint damping and dry friction are set to zero. Identification minimizes the pooled angle RMSE over visible samples in the 0--3~s interval, with equal weighting across load cases. A coarse grid search followed by two local refinements yields $b=52.5~\mathrm{N\,s/m}$ and a fitting RMSE of $7.04^\circ$; evaluation over the reserved 4--7~s interval gives an RMSE of $8.11^\circ$. The fitted $b$ therefore captures the overall damping observed in the joint response rather than an intrinsic material property of the actuator.

\subsection{Mechanical Stop and Joint Behavior}
\label{subsec:joint-stop}

\begin{figure}[!t]
  \centering
  \includegraphics[width=\columnwidth]{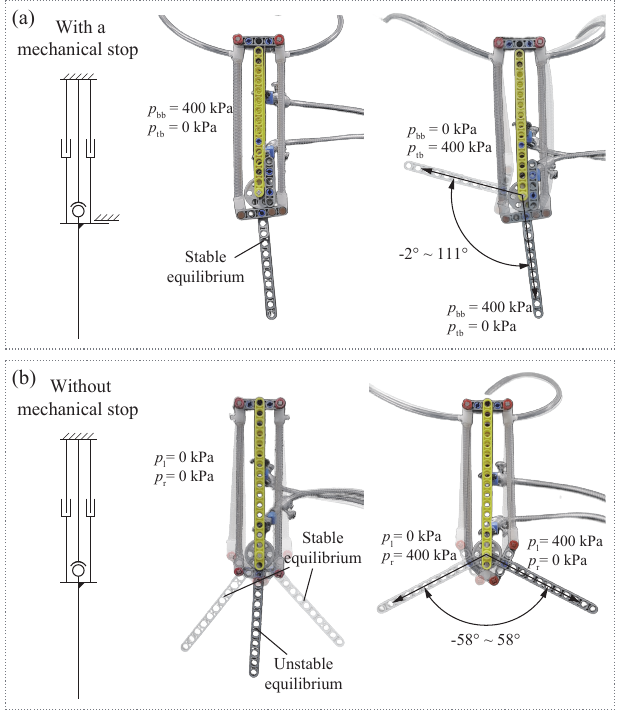}
  \caption{Joint stability and motion range (a) with and (b) without a mechanical stop. The diagrams compare the straight configurations, and the image sequences show the corresponding pressure-limited motion.}
  \label{fig:joint-stop}
\end{figure}

This subsection analyzes the joint stability and motion range to clarify the need for a mechanical stop. The symbols used in this subsection are summarized in Table~\ref{tab:joint-symbols}.

\begin{table}[!t]
  \caption{Notation for joint stability and motion-range analysis.}
  \label{tab:joint-symbols}
  \centering
  \scriptsize
  \setlength{\tabcolsep}{0.06in}
  \renewcommand{\arraystretch}{1.25}
  \begin{tabular}{>{\centering\arraybackslash}m{0.15\columnwidth} m{0.75\columnwidth}}
    \toprule
    Symbol & Definition \\
    \midrule
    $i$; $j$ & Actuator index; pressure-command index \\
    $q$ & Joint angle \\
    $p_i$; $\mathbf p$; $p_{\max}$ & Pressure of actuator $i$; pressure vector; maximum pressure \\
    $\mathbf x_i^{\mathrm f}$; $\mathbf x_i^{\mathrm m}$ & Vectors from the joint axis to actuator $i$'s fixed- and moving-link attachment points, expressed in the corresponding link frames \\
    $\mathbf R(q)$ & Rotation of the moving link from its straight configuration \\
    $r_p$ & Pulley radius \\
    $L_i(q)$; $L_{0,i}$ & Current and rest pneumatic lengths of actuator $i$ \\
    $\ell_i(q)$ & Attachment-to-attachment path length of actuator $i$ \\
    $\varepsilon_i$ & Axial strain of actuator $i$ \\
    $T_i$; $T_{\mathrm{qs}}$ & Actuator tension; quasi-static tension relation \\
    $r_i=-\ell_i'$ & Signed moment arm of actuator $i$ \\
    $k_i$ & Tangent stiffness of actuator $i$ at constant pressure \\
    $a$; $\lambda$ & Actuator-line offset; normalized joint location at $q=0$ \\
    $T_0$; $k$ & Common actuator tension and tangent stiffness at $q=0$ \\
    $\tau$ & Joint torque produced by the actuators \\
    $K_q$ & Tangent joint stiffness \\
    $q^\ast$ & Free-equilibrium angle \\
    \bottomrule
  \end{tabular}
\end{table}

The straight configuration is an important operating posture for a single joint.
The middle column of Fig.~\ref{fig:joint-stop} therefore compares its stability with and without a mechanical stop.
Because the joint with a stop is designed to bend only in the flexion direction, its flexor and extensor actuators have different rest lengths.
The joint without a stop uses equal actuator rest lengths to preserve symmetry.
Under quasi-static conditions and neglecting gravity, friction, and hysteresis, the joint torque produced by the actuators is
\begin{equation}
  \tau(q,\mathbf p)
  =-\sum_i T_i(q,p_i)\ell_i'(q),
  \label{eq:stability-torque}
\end{equation}
where primes denote derivatives with respect to $q$.
Assuming that the actuators remain in tension without slack or buckling, a free equilibrium $q^\ast$ satisfies
\begin{equation}
  \tau(q^\ast,\mathbf p)=0.
  \label{eq:free-equilibrium}
\end{equation}
The tangent joint stiffness at this equilibrium is
\begin{equation}
  K_q=-\left.\frac{\partial\tau}{\partial q}\right|_{q=q^\ast,\mathbf p}
  =\left.\sum_i\left[k_i(\ell_i')^2+T_i\ell_i''\right]
  \right|_{q=q^\ast,\mathbf p},
  \label{eq:stability}
\end{equation}
where $k_i=(\partial T_i/\partial\ell_i)|_{p_i}$. At a free equilibrium, $K_q>0$ gives a locally restoring response, whereas $K_q<0$ implies instability. With a mechanical stop (Fig.~\ref{fig:joint-stop}a), the extension torque presses the moving link against the unilateral constraint. A small force disturbance changes only the constraint reaction while contact is maintained, so the straight posture remains in equilibrium. Without a mechanical stop (Fig.~\ref{fig:joint-stop}b), symmetry makes $q=0$ a free equilibrium if and only if $p_{\mathrm l}=p_{\mathrm r}$. At $q=0$, the actuators are parallel to each other and to the links. Let $a$ be their offset from the joint axis, and $\lambda\ell(0)$ and $(1-\lambda)\ell(0)$ the axial distances from the joint to their fixed and moving attachments. Then $\ell_{\mathrm l}'=-\ell_{\mathrm r}'=a$ and $\ell_{\mathrm l}''=\ell_{\mathrm r}''=-\lambda(1-\lambda)\ell(0)$. With $T_{\mathrm l}=T_{\mathrm r}=T_0$ and $k_{\mathrm l}=k_{\mathrm r}=k$, (\ref{eq:stability}) becomes
\begin{equation}
  K_q(0)
  =2\left[ka^2-T_0\lambda(1-\lambda)\ell(0)\right].
  \label{eq:straight-stiffness}
\end{equation}
The straight configuration is therefore stable only if
\begin{equation}
  \frac{a}{\ell(0)}>
  \sqrt{\frac{T_0}{k\ell(0)}\lambda(1-\lambda)}.
  \label{eq:straight-stability-condition}
\end{equation}
In the constructed joint, the initial extension required to keep both actuators taut over the desired motion range violates this condition, consistent with the perturbation experiment.

The right column of Fig.~\ref{fig:joint-stop} compares the pressure-limited motion ranges. The flexor actuator follows the direct route, whereas the extensor actuator is guided by the pulley:
\begin{equation}
  \begin{aligned}
    \ell_{\mathrm{flex}}(q)
    &=\left\|\mathbf x_{\mathrm{flex}}^{\mathrm f}
    -\mathbf R(q)\mathbf x_{\mathrm{flex}}^{\mathrm m}\right\|,\\
    \ell_{\mathrm{ext}}(q)
    &=\ell_{\mathrm{ext}}(0)+r_pq .
  \end{aligned}
  \label{eq:muscle-length}
\end{equation}
The corresponding strain and tension are
\begin{equation}
  \varepsilon_i(q)
  =\frac{L_i(q)-L_{0,i}}{L_{0,i}},
  \qquad
  T_i(q,p_i)
  =T_{\mathrm{qs}}\!\left(\varepsilon_i(q),p_i\right),
  \label{eq:range-tension}
\end{equation}
Here, $T_{\mathrm{qs}}$ is given by (\ref{eq:quasistatic-force}). Substitution into (\ref{eq:stability-torque}) gives the free joint torque. With $p_{\max}=400$~kPa, the two pressure-limited endpoints are the stable roots under the opposing extreme commands:
\begin{equation}
  \tau\!\left(\hat q_1,(p_{\max},0)\right)=0,
  \qquad
  \tau\!\left(\hat q_2,(0,p_{\max})\right)=0.
  \label{eq:range-endpoints}
\end{equation}
Without a mechanical stop, $q_j=\hat q_j$; with a mechanical stop, $q_j=\max(q_s,\hat q_j)$, where $q_s$ is the lower bound imposed by the stop. The range limits are
\begin{equation}
  q_{\min}=\min_j q_j,
  \qquad
  q_{\max}=\max_j q_j.
  \label{eq:range-stop}
\end{equation}
Substitution gives approximately $-58^\circ$ to $58^\circ$ without the stop and $-2^\circ$ to $111^\circ$ with it, consistent with the experimental results in Fig.~\ref{fig:joint-stop}. Although the total spans are similar, the stop shifts nearly all available rotation to the intended flexion side. The mechanical stop is therefore necessary both to stabilize the straight posture and to prevent the pressure-limited range from extending into unintended hyperextension. The same stop is incorporated into the finger modules described in Section~\ref{subsec:hand}.

\section{Reconfigurable Robot Examples}
\label{sec:robots}

\subsection{Robotic Hand}
\label{subsec:hand}

\begin{figure*}[!t]
  \centering
  \includegraphics[width=\textwidth]{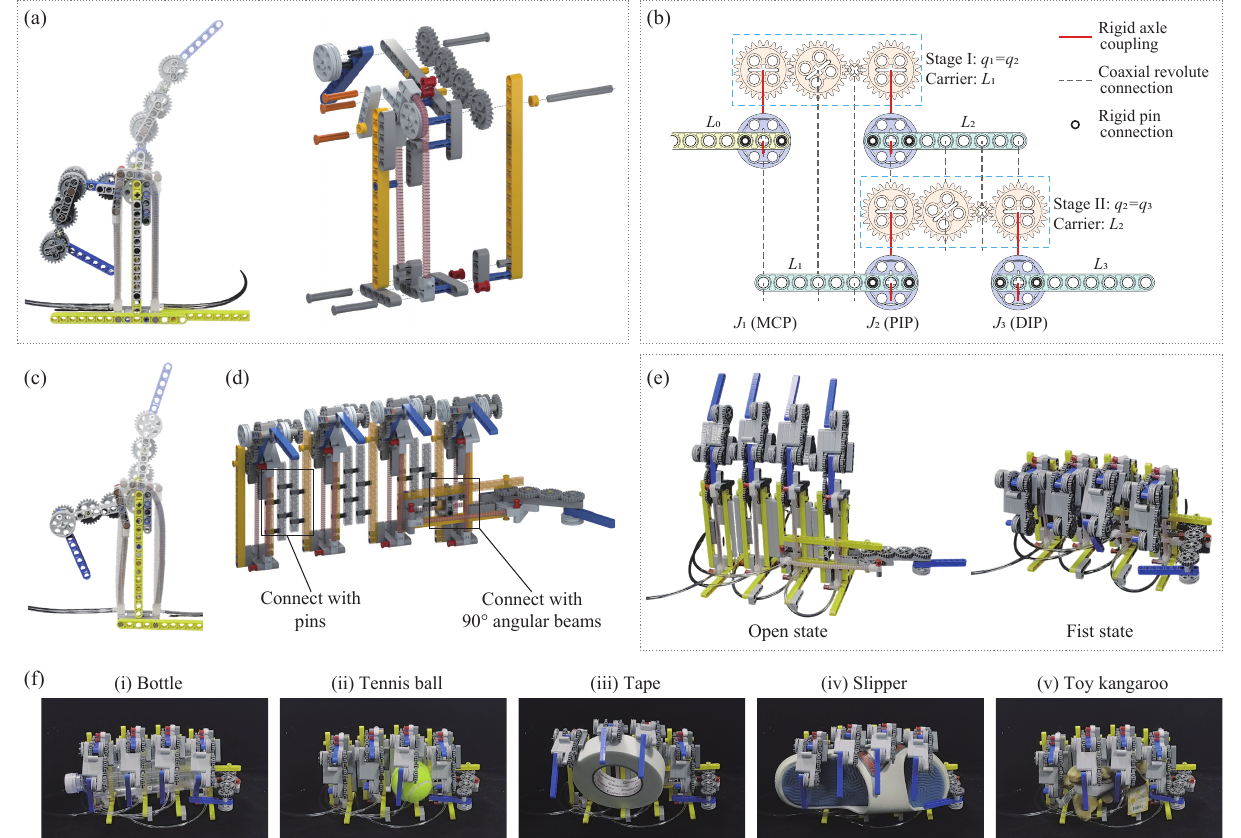}
  \caption{Robotic hand constructed using MUSBRIC. (a) Three-joint finger module and exploded assembly. (b) Two $1{:}1$ gear stages enforcing equal relative rotations of the three joints. (c) Two-joint thumb constructed from the first two joints of the finger module with one gear stage. (d) Assembly of four fingers and the thumb. (e) Open and fist states. (f) Grasps of a bottle, tennis ball, tape roll, slipper, and toy kangaroo.}
  \label{fig:hand}
\end{figure*}

We extend the single-joint mechanism into two digit modules: a three-joint finger and a two-joint thumb. In each three-joint finger, an antagonistic actuator pair drives the proximal joint (Fig.~\ref{fig:hand}a). Two successive $1{:}1$ gear stages transmit this rotation and enforce equal relative rotations $q_1=q_2=q_3$ (Fig.~\ref{fig:hand}b). The two-joint thumb reuses the first two joints of this design and therefore requires only one gear stage (Fig.~\ref{fig:hand}c). Both modules incorporate the mechanical stops analyzed in Section~\ref{subsec:joint-stop} to retain a stable extended posture.

The five digit modules are joined with pins and $90^\circ$ angular beams to form a robotic hand (Fig.~\ref{fig:hand}d). Alternating open-loop pressure commands switch the assembled hand between open and fist states (Fig.~\ref{fig:hand}e). The compliant actuators allow the fingers to adapt passively during contact. The hand grasps a bottle, tennis ball, tape roll, slipper, and a soft plush kangaroo, spanning a range of sizes, shapes, and stiffnesses. These results demonstrate the integration of compliant actuation with standard LEGO structures and transmissions.

\subsection{Leg and Quadruped}
\label{subsec:leg}

\begin{figure*}[!t]
  \centering
  \includegraphics[width=\textwidth]{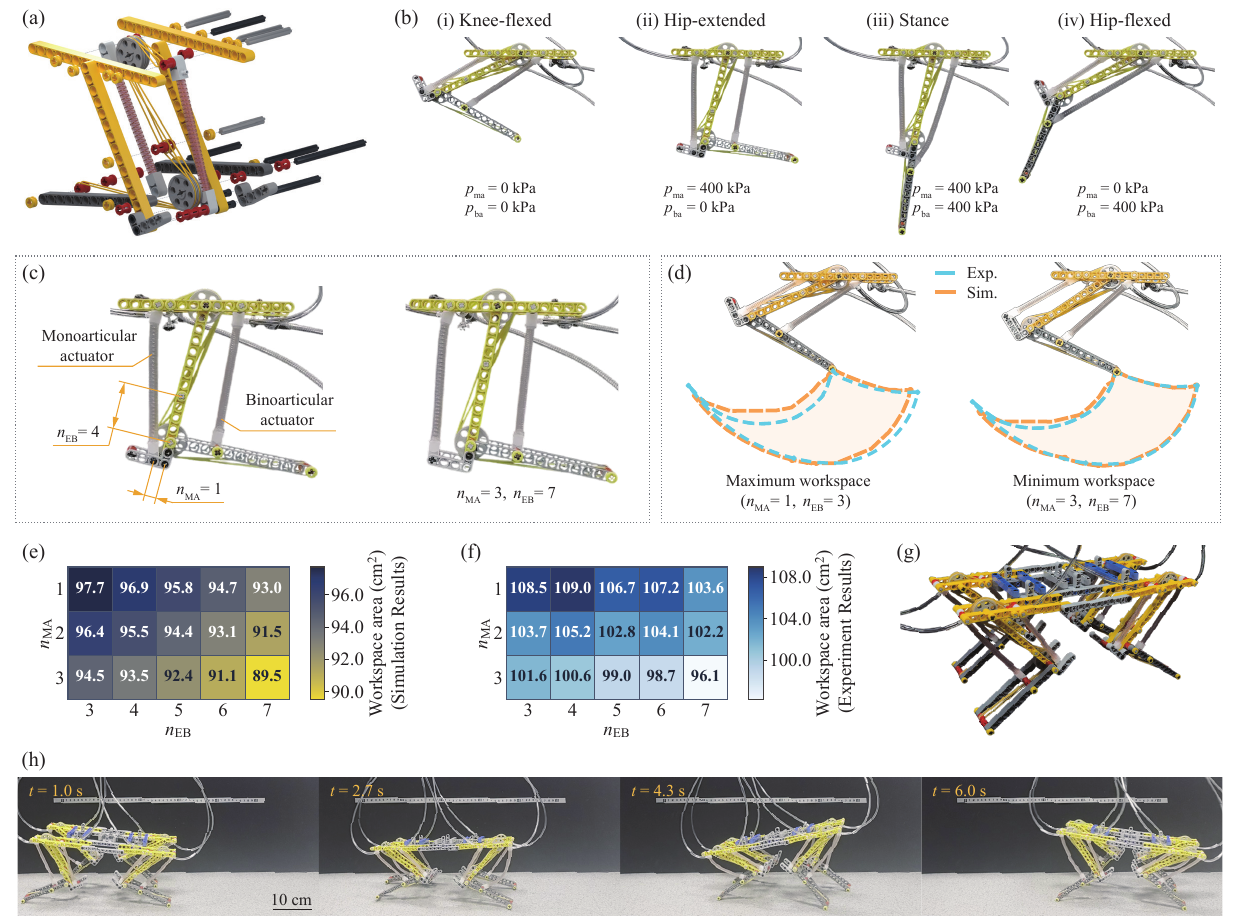}
  \caption{Leg reconfiguration and quadruped locomotion. (a) Exploded assembly of the two-joint leg. (b) Four-posture motion cycle of the leg. (c) LEGO-grid attachment parameters and representative leg configurations. (d) Configurations with the maximum and minimum workspace areas and their corresponding workspaces. (e) Simulated and (f) experimental parameter maps of the workspace area across the 15 configurations. (g) A quadruped robot assembled from four leg modules.
  (h) Open-loop walking sequence of the quadruped assembled from four leg modules; the robot reaches 0.21 body lengths per second.}
  \label{fig:leg}
\end{figure*}

We assemble a reconfigurable two-joint leg for geometry exploration and use four such modules to construct a quadruped robot. Following a previously reported musculoskeletal bionic leg~\cite{dong2025bionicleg}, the leg comprises hip and knee joints driven by monoarticular and biarticular actuators (Fig.~\ref{fig:leg}a). The monoarticular actuator spans only the hip, whereas the biarticular actuator spans both joints. To simplify the pneumatic system, two elastic bands provide antagonistic restoring forces in place of additional actuators, with one band spanning each joint. The four commands $(p_{\mathrm{ma}},p_{\mathrm{ba}})=(0,0),(400,0),(400,400),(0,400)$~kPa generate a motion cycle of the leg (Fig.~\ref{fig:leg}b).

A leg containing four elastic elements admits many choices for their attachment points, which determine the actuator routing and foot motion. Scanning these geometric parameters in a conventional soft--rigid hybrid robot requires fabricating multiple structural variants. With MUSBRIC, the attachment points can instead be relocated on the LEGO grid in both the physical assembly and its simulation model. We demonstrate this capability by varying the LEGO-grid offsets of the monoarticular actuator and elastic-band attachment points, denoted by $n_{\mathrm{MA}}\in\{1,2,3\}$ and $n_{\mathrm{EB}}\in\{3,4,5,6,7\}$, respectively (Fig.~\ref{fig:leg}c). Each of the resulting 15 configurations is assembled physically and modeled in simulation. Experimental foot trajectories are obtained by tracking the foot endpoint in video, whereas simulated trajectories are calculated directly by the simulator. The resulting foot workspaces are illustrated in Fig.~\ref{fig:leg}d, and their area $A(n_{\mathrm{MA}},n_{\mathrm{EB}})$ provides a common metric for the simulated and experimental parameter maps in Fig.~\ref{fig:leg}e and f. These workspaces represent motion under the sampled commands rather than every kinematically reachable foot position.

All 15 configurations use the same leg model, differing only in the attachment coordinates, with the upper frame fixed as in the experiments. Both pneumatic actuators remain straight and are represented by the axial model in Section~\ref{subsec:characterization}. Each elastic band is modeled using its measured force--strain relation. The experimental pressure sequence is applied in 1~s steps. 
Each configuration is initialized from the same posture and simulated for two 4~s cycles; its workspace area is computed from the foot trajectory in the second cycle.

The simulated workspace area decreases monotonically with both $n_{\mathrm{MA}}$ and $n_{\mathrm{EB}}$, from 97.7~cm$^2$ at $(1,3)$ to 89.5~cm$^2$ at $(3,7)$ (Fig.~\ref{fig:leg}e). The experiments show the same overall dependence on attachment geometry: increasing $n_{\mathrm{MA}}$ consistently reduces the area, while increasing $n_{\mathrm{EB}}$ produces a general decrease with small local variations (Fig.~\ref{fig:leg}f). The measured areas at $(1,3)$ and $(1,4)$ are 108.5 and 109.0~cm$^2$, respectively. This 0.5~cm$^2$ difference is treated as experimental variation, and $(1,3)$ is taken as the maximum-workspace configuration; $(3,7)$ gives the minimum measured area of 96.1~cm$^2$. Thus, the simulation captures the observed dependence on attachment geometry and identifies the same extrema, although it underestimates the absolute area.

Finally, four leg modules are attached symmetrically to a LEGO body to form the quadruped in Fig.~\ref{fig:leg}g. The robot uses an open-loop four-beat gait. Because the front and rear leg pairs are mounted in mirrored orientations, the rear legs execute the pressure cycle in the reverse order of the front legs, while the four leg commands are evenly phase-shifted by one quarter cycle ($90^\circ$). The quadruped reaches a forward speed of 0.21 body lengths per second, demonstrating that the same component set scales from a reconfigurable leg to a functional walking robot (Fig.~\ref{fig:leg}h).

\section{Discussion and Conclusion}
\label{sec:conclusion}

This paper presented MUSBRIC, a modular hardware platform and matched simulator for reconfigurable musculoskeletal robots. LEGO-compatible endcaps turn fiber-reinforced extensile actuators into reusable muscle bricks that connect directly to standard LEGO components. The identified actuator relation is incorporated into MuJoCo, enabling the same attachment geometry to be evaluated in simulation.

The single-joint study validates flexion--extension motion under pressure steps and shows how actuator routing and joint constraints determine behavior. Pulley-guided routing maintains the extensor moment arm, while the mechanical stop stabilizes the straight posture and shifts the motion range toward flexion. The hand extends this mechanism into a compliant multi-joint structure capable of grasping varied objects. The leg study demonstrates parameter scanning across 15 matched physical and simulated configurations, and four leg modules form a walking quadruped.

The current actuator model is quasi-static with an added damping term, the hand has limited independently controlled degrees of freedom, and the quadruped has been tested with one gait. Future work will improve dynamic modeling, increase hand controllability, and evaluate more demanding locomotion.

\balance
\bibliographystyle{IEEEtran}
\bibliography{references}

\end{document}